%% file: cvpr2014.tex
\documentclass[10pt,twocolumn,letterpaper]{article}

\usepackage{cvpr}
\usepackage{times}
\usepackage{graphicx}
\usepackage{amsmath}
\usepackage{amssymb}
\usepackage{subfigure}
\usepackage{caption}
\usepackage{multirow}

\usepackage{stmaryrd}
\usepackage{xspace}

\usepackage{algorithm}
\usepackage{algorithmic}
\usepackage{placeins}

\makeatletter

\DeclareRobustCommand\onedot{\futurelet\@let@token\@onedot}
\def\@onedot{\ifx\@let@token.\else.\null\fi\xspace}

\def\iid{{i.i.d}\onedot}
\def\eg{{e.g}\onedot} 
\def\ie{{i.e}\onedot}

\makeatother

\DeclareSymbolFont{bbold}{U}{bbold}{m}{n}
\DeclareSymbolFontAlphabet{\mathbbold}{bbold}
\newcommand{\ind}[1]{\llbracket{#1}\rrbracket}
\usepackage[pagebackref=true,breaklinks=true,letterpaper=true,colorlinks,bookmarks=false]{hyperref}

\cvprfinalcopy 

\ifcvprfinal\fi
\begin{document}
\addtolength{\dbltextfloatsep}{-\baselineskip}
\addtolength{\textfloatsep}{-.9\baselineskip}

\title{Closed-Form Training of Conditional Random Fields \\for Large Scale Image Segmentation}

\author{
Alexander Kolesnikov \\
IST Austria \\
\texttt{akolesnikov@ist.ac.at} \\
\and
Matthieu Guillaumin\\
ETH Z\"urich\\
\texttt{guillaumin@vision.ee.ethz.ch} \\
\and
\hspace*{-24mm} Vittorio Ferrari\\
\hspace*{-24mm} University of Edinburgh \\
\hspace*{-24mm} \texttt{vittorio.ferrari@ed.ac.uk} \\
\and
\hspace*{10mm} Christoph H. Lampert\\
\hspace*{10mm} IST Austria \\
\hspace*{10mm} \texttt{chl@ist.ac.at} \\
}

\maketitle

\newcommand{\METHOD}{LS-CRF}

\begin{abstract}
We present \METHOD{}, a new method for very efficient large-scale  
training of Conditional Random Fields (CRFs). 
It is inspired by existing closed-form expressions for the maximum 
likelihood parameters of a generative graphical model with 
tree topology. 
\METHOD{} training requires only solving a set of independent regression problems, 
for which closed-form expression as well as efficient iterative solvers are
available.
This makes it orders of magnitude faster than conventional maximum likelihood 
learning for CRFs that require repeated runs of probabilistic inference. 
At the same time, the models learned by our method still allow for joint inference at test time. 

We apply \METHOD{} to the task of semantic image segmentation, 
showing that it is highly efficient, even for loopy models where 
probabilistic inference is problematic. 
It allows the training of image segmentation models from significantly 
larger training sets than had been used previously. 
We demonstrate this on two new datasets that form a second contribution 
of this paper. They consist of over 180,000 images with figure-ground 
segmentation annotations. 
%
%
Our large-scale experiments show that the possibilities of CRF-based 
image segmentation are far from exhausted, indicating, for example, 
that semi-supervised learning and the use of non-linear predictors are 
promising directions for achieving higher segmentation accuracy in the future.
%
\end{abstract}

%
%
%

\input{introduction}
\input{method}

\begin{figure*}[t]
    \centering
    \includegraphics[width=.85\textwidth]{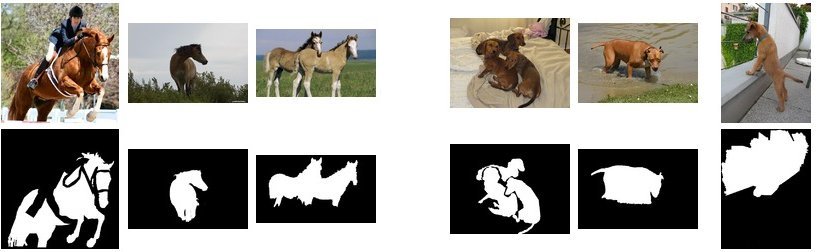}
    \caption{Example images and segmentation masks from \emph{HorseSeg} (left) and \emph{DogSeg} (right) datasets. 
    For each dataset, we illustrate: manually created annotation (left), annotation from bound boxes (middle) 
    and annotation from per-image labels (right). Annotation from bounding boxes is typically rather accurate, 
    annotation from labels can contain significant errors.}
    \label{fig:examples}
\end{figure*}

\input{relatedwork}

\input{experiments}

\input{summary}
\renewcommand{\baselinestretch}{.93}
{\small
\bibliographystyle{ieee}
\bibliography{cvpr2014}
}

\clearpage

\input{supplement}

\end{document}

%% file: introduction.tex
\begin{figure*}[t]
\centering
\subfigure[superpixel graph]{
\quad\includegraphics[height=.2\textwidth]{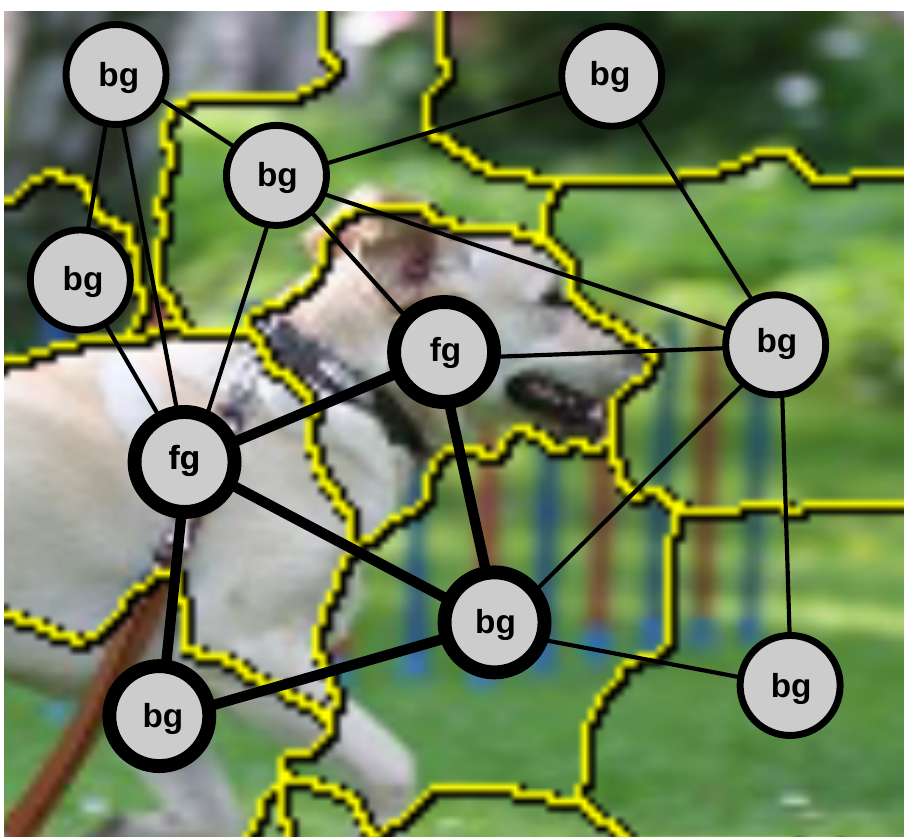}\quad
}\quad
\subfigure[pairwise decomposition]{
\qquad\includegraphics[height=.2\textwidth]{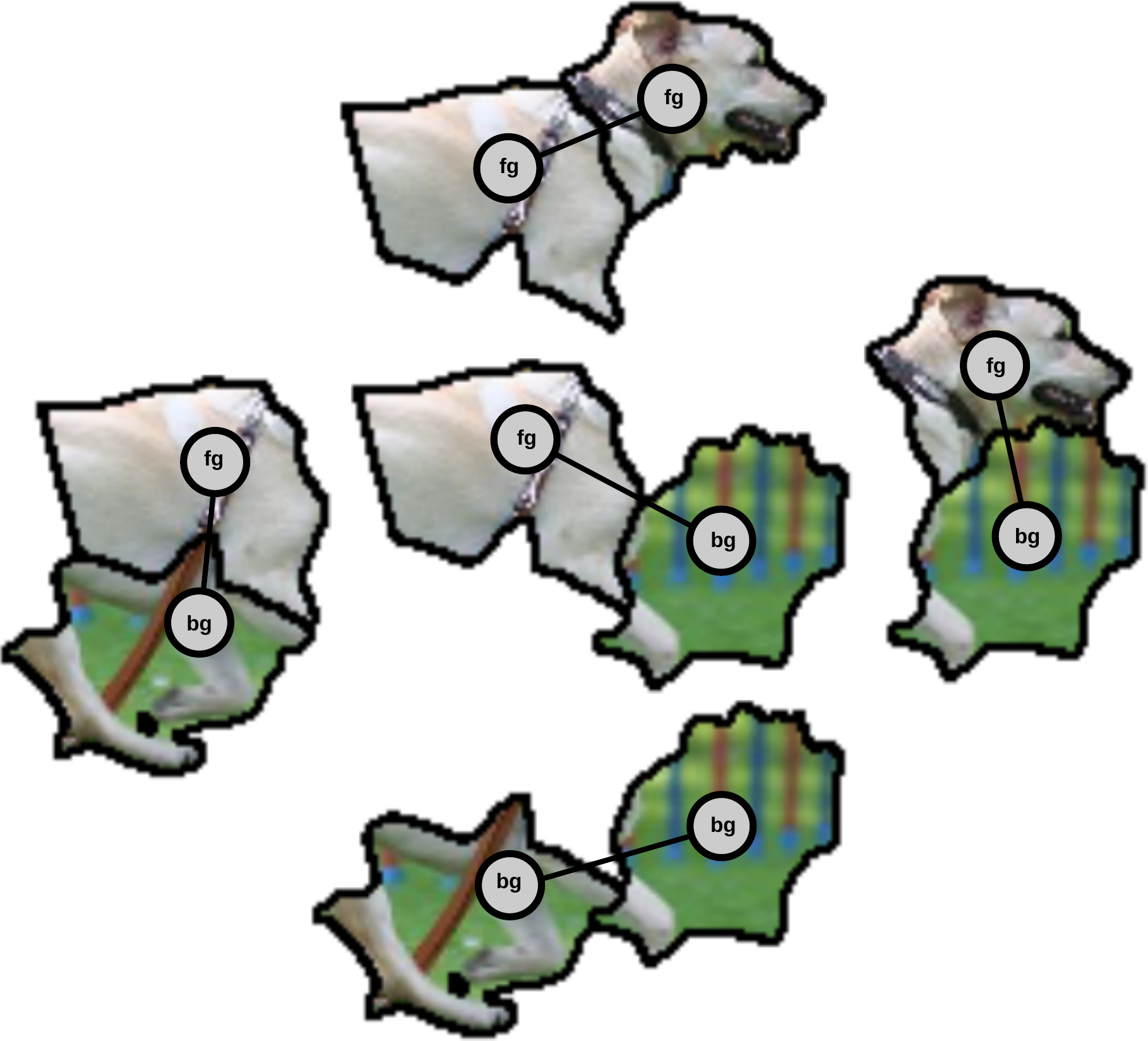}\qquad
}\quad
\subfigure[parameter regression]{ 
\quad\includegraphics[height=.2\textwidth]{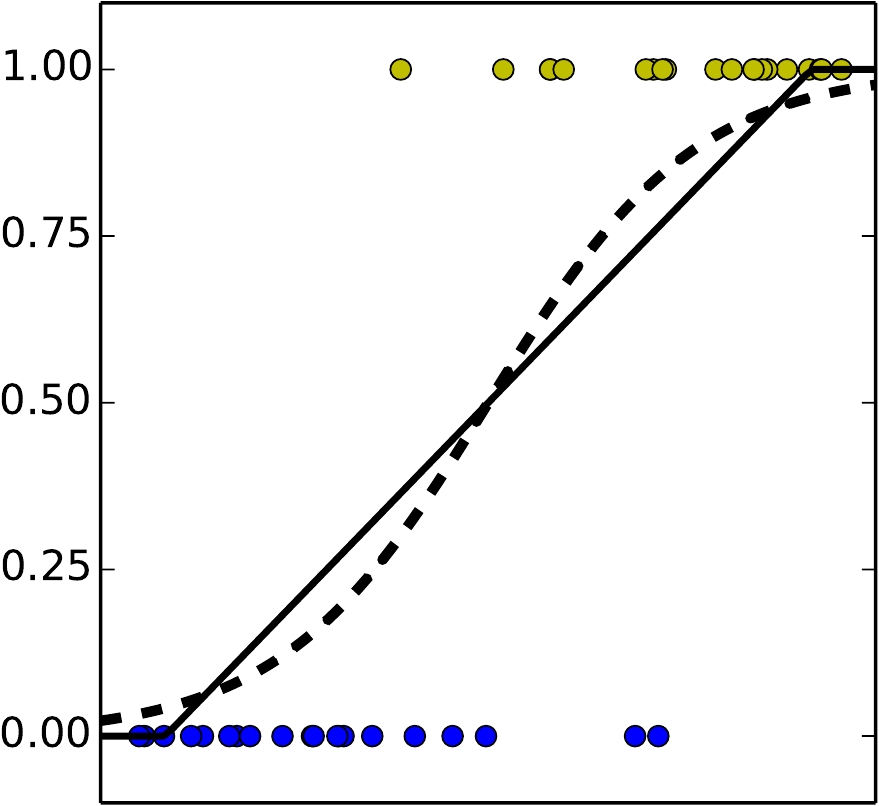}\quad
}
\caption{Schematic illustration of \METHOD{}  for image segmentation -- training phase: 
(a) we are given images with predefined graph structure (here based on superpixels) 
and per-node annotation (here: {\bf\textsf{fg/bg}}), (b) we form training 
subproblems from all edges in the graph (shown for bold subgraph), (c) for each label 
combination, we train a linear (solid line) or nonlinear (dashed line) regressor to 
predict the label combination's conditional probability.}\label{tab:LSCRF-train}
\subfigure[predicted energy]{
\quad\includegraphics[height=.2\textwidth]{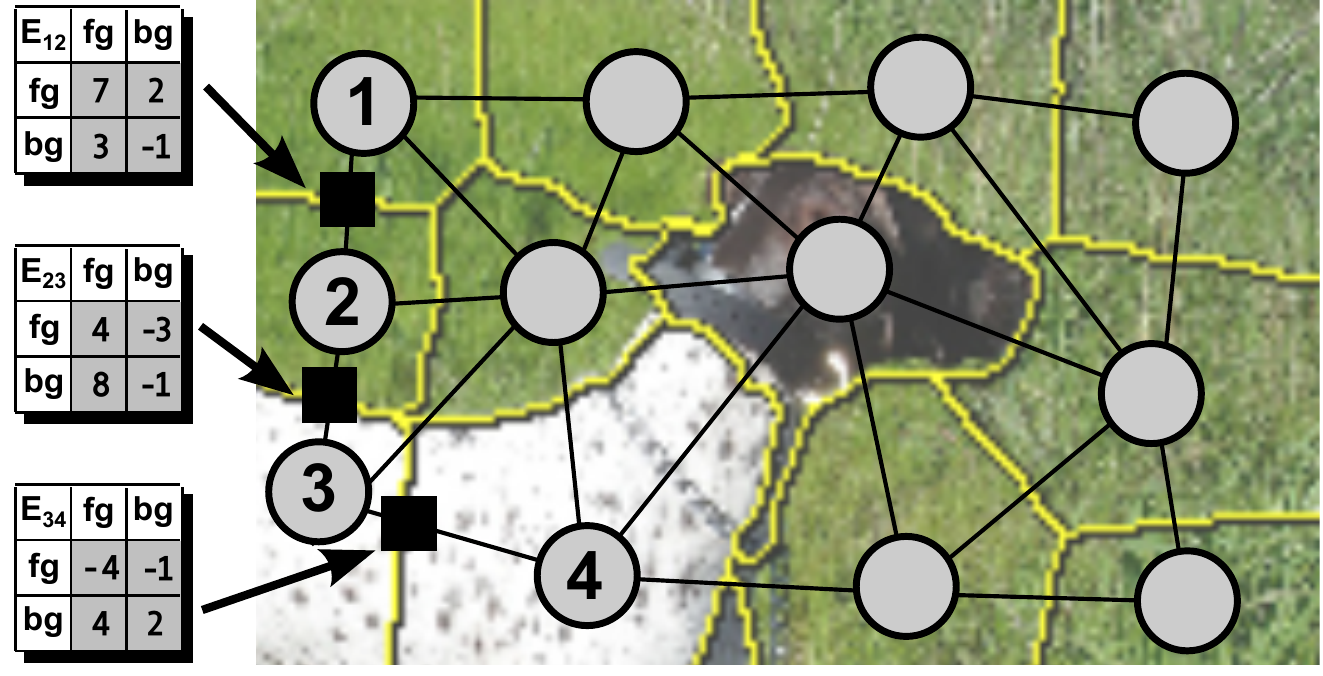}\quad
}\quad
\subfigure[segmentation result]{
\quad\includegraphics[height=.2\textwidth]{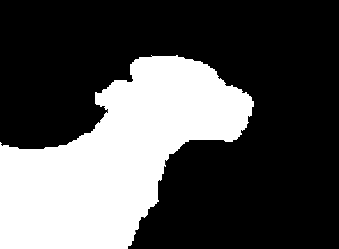}\quad
}
\caption{Schematic illustration of \METHOD{}  for image segmentation -- prediction phase: 
(a) for a new image we use the regressors to predict an energy table for each pairwise term 
(visualized on three edges between four nodes),
(b) The energy function defines a probability distribution which yields a segmentation 
(here: by MAP prediction).}\label{tab:LSCRF-test}
\end{figure*}

\section{Introduction}\label{sec:intro}
Conditional random fields (CRFs)~\cite{LaffertyICML01} have 
emerged as powerful tools for modeling several interacting 
objects, such as parts of bodies, or pixels in an image. 
As a consequence, they have found multiple applications in 
computer vision, in particular in \emph{human pose estimation}, 
\emph{action recognition} and \emph{semantic image segmentation}~\cite{nowozin2011structured}.
However, the increased modeling capacity of conditional random 
fields comes at a computational price: 
exact training of CRFs requires repeated runs of probabilistic inference. 
For simple chain- or tree-structured models this is possible but 
computationally expensive. For loopy models, as they are 
dominant in image segmentation, efficient exact algorithms 
are provably intractable\footnote{See Section~\ref{sec:relatedwork} 
for the exact characterization of this statement.}. 
Therefore, an important problem in machine learning and also
computer vision research is to find methods for fast \emph{approximate} 
training of loopy CRFs. 

In this work we propose a new technique for fast approximate 
training of CRFs with up to pairwise terms. We call it 
\METHOD{}, where the LS stands both for 
\emph{least squares} and \emph{large scale}. 
We derive \METHOD{} in Section~\ref{sec:method} from existing 
closed-form expressions for the maximum likelihood parameters 
of a non-conditional probability distribution when the 
underlying graphical model is tree-shaped.
We adapt these to the situation of a conditional probability 
distributions modelled by a CRF, obtaining a training problem 
that requires only solving several suitably constructed 
regression problems.
For the most common linear parametrization of the CRF energy, 
these can be solved efficiently using either classical 
closed-form expressions, or using iterative numerical 
techniques, such as the conjugate gradient method, or
simple stochastic gradient descent. 
However, we can also opt for a non-linear parameterization, 
leading to increased expressive power while still staying
computationally feasible. 
Overall, we obtain a CRF training algorithm that is 

\begin{itemize}

\item \textbf{efficient \& scalable} (training sets can have 100,000 images 
or more, subproblems can be solved independently, 
each subproblem can be parallelized itself)

\item \textbf{flexible} (we can, \eg, incorporate non-linearities, 
per-sample weights, or compensate for class imbalance), 

\item \textbf{easy to implement} (\eg, the closed-form expressions 
can be solved in a few lines of Matlab or Python code).
\end{itemize}

While \METHOD{} is a general-purpose training method, it is 
particularly suitable for the type of CRFs that occur in 
computer vision applications, such as image segmentation, where: 
1) the underlying graph is loopy,
2) all variables are of the same "type" (\eg pixels or superpixels),
3) each variable takes values in a rather small label set, 
4) many training examples are available. 

Figures~\ref{tab:LSCRF-train} and~\ref{tab:LSCRF-test} 
illustrate \METHOD{} for semantic image segmentation. 
A formal definition and justification are given in Section~\ref{sec:method}.
Our contribution lies in the steps \ref{tab:LSCRF-train}(b), 
\ref{tab:LSCRF-train}(c) and \ref{tab:LSCRF-test}(a),
which we explain in detail in Section~\ref{sec:method}. 

We demonstrate the usefulness of our method by applying it to 
several image segmentation tasks, both in a semantic multi-class 
setup as well as a binary figure--ground situation.
Our goal in this is to pave the road towards truly large-scale 
experiments in the area of image segmentation. 
Since current datasets rarely consists of more than a few hundred 
images, we created two new datasets of together over 180,000 
images. Some of these have manually created annotation, for 
others, we provide segmentation masks created in a semi-automatic 
way. 
By making these publicly available we hope to stimulate more research 
in two directions: large-scale CRF training and large-scale 
(potentially semi-supervised) image segmentation.

We report results of \METHOD{} in Section~\ref{sec:experiments} on 
this dataset as well as on the Stanford backgrounds dataset, comparing 
our results to baseline techniques in terms of training speed as well 
as segmentation accuracy. 
%

%% file: method.tex
\section{Inference-Free CRF Training}\label{sec:method}
We first reiterate the classical construction of maximum likelihood 
parameter estimation for non-conditional probability distribution 
with tree-shaped independence relations (see \cite[page 150]{wainwright2008graphical}).
For $m$ variables, $y_1,\dots,y_m$, where each $y_s$ can take 
values in label set $\mathcal{L}=\{1,\dots,r\}$, let 
$\mathcal{Y}$ be the set of all possible joint labelings, 
\ie $\mathcal{Y}=\mathcal{L}^m$, and let 
$P_\theta(y)$ for $y=(y_1,\dots,y_m)$ be a joint 
probability distribution over such labelings.
%
We assume the structure of an undirected graphical model 
with underlying graph $G = (V, \mathcal{E})$ with $V=\{1,\dots,m\}$ 
and edge set $\mathcal{E}\subset V\times V$, such that 
$P_\theta(y) \propto \exp(-E_\theta(y))$ for an energy function
\begin{align}
E_\theta(y) &= \sum_{s \in V} \sum_{j\in\mathcal{L}}\theta_{s;j} 
\ind{y_s\!=\!j}  
\\ &\quad + \sum_{(s,t) \in \mathcal{E}}\  \sum_{(j,k)\in\mathcal{L}\times\mathcal{L}} 
\theta_{st;jk}\ind{y_s\!=\!j\wedge y_t\!=\!k },\notag
\end{align}
where $\ind{ P}=1$ if the predicate $P$ is true, and $0$ otherwise.

The distribution $P_\theta$ is fully determined by its 
parameter vector $\theta$ that consists of the unary and 
pairwise parameters $\theta_{s;j}\in\mathbb{R}$ for all $s\in V$ and $j\in\mathcal{L}$, 
and $\theta_{st;jk}\in\mathbb{R}$ for all $(s,t)\in \mathcal{E}$ 
and $(j,k)\in\mathcal{L}\times\mathcal{L}$, respectively.

To \emph{learn} such a distribution means to estimate an unknown $\theta$ 
from a set of \iid samples $(y^{1},\dots,y^{T})$ using the maximum likelihood 
principle. 
It is a classical but rarely used fact that for tree-shaped graphs this 
step is possible in closed form. 
Let 
\begin{align}
  \hat \mu_{s;j} &= \frac{1}{T}\!\sum_{i=1}^{T} \ind{ y_{s}^{i}\!=\!j}, 
  \quad\hat \mu_{st;jk} = \frac{1}{T}\! \sum_{i=1}^{T}\ind{y_{s}^{i}\!=\!j \wedge y_{t}^{i}\!=\!k},  \label{eq:mu_sj}
\end{align}
be the empirical estimates of the single and pairwise variable marginal probabilities 
for all $s\in V$ and $j\in \mathcal{L}$, and for all $(s,t)\in \mathcal{E}$ and $(j,k)\in\mathcal{L}\times\mathcal{L}$.
%
Then 
\begin{align}
\hat \theta_{s;j}   &= \log \hat \mu_{s;j} \label{eq:theta_sj},
\qquad
\hat \theta_{st;jk} = \log \frac{\hat \mu_{st;jk}}{\hat \mu_{s;j} \hat \mu_{t;k}} 
\end{align}
maximum likelihood parameter estimate, as can be checked by a straight-forward computation~\cite{wainwright2008graphical}.
This estimate is \emph{consistent}, \ie $P_{\hat\theta}(y)$ converges to $P_{\theta}(y)$ for $T\to\infty$. 

%

\subsection{Conditional Distributions (\METHOD{} Training)}
The goal of this work is to generalize the above closed-form expression 
into a procedure for inference-free conditional random field (CRF) learning. 
The main problem we have to solve is that above section applies only to 
ordinary fixed probability distribution, whereas a CRFs encode conditional 
distributions, \ie a different distribution $P(y|x)$ for each $x$ from a 
set $\mathcal{X}$.
%

We achieve this data-dependence by making the parameters $\theta_{s;j}$ 
and $\theta_{st;jk}$ functions of $x$ instead of constants, so 
\begin{align}
    P_{\theta}(y|x) & = P_{\theta(x)}(y) \propto \exp(-E_{\theta(x)}(y)) \label{eq:dist}
\intertext{with}
E_{\theta(x)}(y) &= \sum_{s \in V} \sum_{j\in\mathcal{L}}\theta_{s;j}(x)\ind{ y_s\!=\!j}
\label{eq:energy}
\\ &\quad+ \sum_{(s,t) \in \mathcal{E}} \sum_{(j,k)\in\mathcal{L}\times\mathcal{L}} \theta_{st;jk}(x)\ind{ y_s\!=\!j\wedge y_t\!=\!k }. \notag
\end{align}

Given a training set, $\{(x^1,y^1), (x^2, y^2), \dots, (x^T,y^T)\}$, of \iid 
examples from an unknown distribution $d(x,y)$ 
this leads to the following learning problem:
identify functions $\hat\theta_{s;j}:\mathcal{X}\to\mathbb{R}$ 
for $s\in V$ and $j\in\mathcal{L}$, as well as ${\hat\theta_{st;jk}:\mathcal{X}\to\mathbb{R}}$
for $(s,t)\in\mathcal{E}$ and $(j,k)\in\mathcal{L}\times \mathcal{L}$, such that for every $x^i$, 
$P_{\hat\theta(x^i)}(y)$ is an as good as possible estimate of the unknown conditional distribution $d(y|x^i)$. 

Inspired by the data-independent case, we again first define the unary and pairwise 
marginals of the training examples. Since there is only one labeling of each training example, these are 
just indicators of the occurring labels,
\begin{align}
\hat\mu_{s;j}(x^i) &= \ind{ y^i_s\!=\!j},
\quad\hat\mu_{st;jk}(x^i) = \ind{ y^i_s\!=\!j \wedge y^i_t\!=\!k}
\label{eq:learn_mu_st}
\end{align}
for $i=1,\dots,T$. We then learn (regression) functions 
\begin{align}
f_{s;j}:\mathcal{X}\to (0,1)\subset\mathbb{R},  \quad\ f_{st;jk}:\mathcal{X}\to (0,1)\subset \mathbb{R},
\label{eq:functions}
\end{align}
that generalize these marginals, i.e. they approximately reproduce the values 
observed on the training data, but generalize to the complete domain $\mathcal{X}$.
Subsequently, we continue analogously to the data-independent situation.
We set 
\begin{align}
\hat \theta_{s;j}(x)\!=\!\log\!f_{s;j}(x), \ \ \ 
\hat \theta_{st;jk}(x)\!=\!\log\!\frac{f_{st;jk}(x)}{f_{s;j}(x)\,f_{t;k}(x)},
\label{eq:theta_f_stjk}
\end{align}
from which we obtain the conditional CRF distribution using Equation~\eqref{eq:dist} and \eqref{eq:energy}.

Overall, we have reduced the problem of maximum likelihood CRF learning to 
problem of learning a set of independent regression functions, one per edge 
(or edge type) and label combination (Algorithm~\ref{alg:LSCRF-training}). 
Note that for any node, $s$, that is part of at least one edge, $(s,t)$, we do 
not have to learn the unary functions, $f_{s;j}(x)$ by a regression steps. 
Instead, the marginalization condition, $\hat\mu_{s;j}(x^i)=\sum_{k\in\mathcal{L}}\hat\mu_{st;jk}(x^i)$, 
implies that we can obtain it from the pairwise functions as $f_{s;j}(x)=\sum_{k\in\mathcal{L}} f_{st;jk}(x)$.
Thus, we train unary regression functions only for isolated nodes.

The main advantage of \METHOD{} to classical maximum-likelihood or maximum-margin 
CRF training is its efficiency, since no probabilistic inference or energy minimization 
routines must be run over the training data in order to estimate parameters. 
Also, the regression problems rely only on observed data and have no dependencies 
between them, so they can be solved completely in parallel or even in a distributed way. 
The outcome, however, are not just independent per-variable score, but a proper 
energy function including pairwise terms that we can use for joint predictions 
by probabilistic inference or MAP prediction. 

\begin{algorithm}[t]
\begin{algorithmic}[1]
\INPUT training data: $(x^i,y^i,G^i)_{i\in I}$ \ for $I=\{1,\dots,n\}$,

$x^i$: images, $y^i$: ground truth, $G^i=(V^i,\mathcal{E}^i)$: graphs\medskip

\STATE set $\phi^i_{st}\in\mathbb{R}^{D}\ \ \leftarrow$\ \ feature vector of edge $(s,t)\in \mathcal{E}^i$
\FOR{$j,k\in\mathcal{L}\!\times\!\mathcal{L}$}
\STATE set $\mu^i_{st;jk} \leftarrow \ind{ y^i_s\!=\!j \wedge y^i_t\!=\!k}\ \quad\forall i\in I,(s,t)\in \mathcal{E}^i$
\STATE train $f_{jk}$ \ \ from\ \ $\bigcup_{i\in I}\bigcup_{(s,t)\in\mathcal{E}^i} \big\{(\phi^i_{st},\mu^i_{st;jk})\big\}$
\ENDFOR\medskip
\OUTPUT functions $f_{jk}\!:\mathbb{R}^D\to (0,1)$\ \ for\ \ $j,k\in\mathcal{L}\!\times\!\mathcal{L}$ 
%
\end{algorithmic}
\caption{\METHOD{} -- Training}\label{alg:LSCRF-training}
\end{algorithm}

\subsection{Parameterization}
We illustrate two setup for learning the regression functions, \eqref{eq:functions}, in practice: 
linear least-squares regression, and non-linear regression by gradient boosted decision trees.

To improve the readability, we drop the indices indicating the label pair, $jk$, 
from the notation, understanding that all following steps have to be performed 
separately for each $(j,k)\in\mathcal{L}\times\mathcal{L}$.
We furthermore assume, also for the sake of a more compact notation, that 
only one class of pairwise factors occur. This allows us to also drop the edge 
identifier, $st$, from the notation.
As a consequence, multiple regions within each image will contribute to 
learning the same pairwise term. We write $\phi^1,\dots,\phi^N$ for the 
feature representations of all such regions across all images, and 
$\mu^1,\dots,\mu^N$ for their estimated marginals, \eqref{eq:learn_mu_st}, 
obtained from the ground truth annotation. 
Note that $N$ is typically in the order of the total number of edges the 
graphs of all training images, so much larger than the number 
of training examples, $n$. 

\paragraph{Linear Models.} 
In the linear case, we parameterize\footnote{We suppress a possible bias term in 
the regression. It can be recovered by adding an additional, constant, entry to 
the feature representation.} $f(x)=\langle w,\phi(x)\rangle$ 
and use least squares regression to obtain the weight vector, $w$, 
\begin{align}\label{eq:ridge-regression}
\operatorname{min}_{w}   \quad \lambda \|w\|^2 + \sum_{i=1}^N \| f(x^i) - \hat\mu(x^i)  \|^2,
\end{align}
where $\lambda\geq 0$ is an (optional) regularization constant. 
A major advantage of this formulation is that 
Equation~\eqref{eq:ridge-regression} has a closed form solution for the optimal weight vector,
\begin{align}\label{eq:closed-form}
w = (\Phi\Phi^\top + \lambda \operatorname{I})^{-1}\Phi \boldsymbol{\mu}
\end{align}
where $\Phi$ is the matrix with the features, $\phi^1,\dots\,\phi^N$,
as columns, and $\boldsymbol{\mu}$ is the vector formed from the target outputs, $\mu^1,\dots,\mu^N$. 

Computing Equation~\eqref{eq:closed-form} for $D$-dimensional feature vectors
requires solving a linear system of size $D\times D$.
This is possible efficiently even for features that have several thousands dimensions,
and using a precomputed LR-factorization of the matrix $(\Phi\Phi^\top\!\!\!+\!\lambda \operatorname{I})$,
it is even possible to solve the regression problems for all label pair with minimal overhead 
compared to the time for a single pair.
When the number of training examples is very large, however, it can happen that the computation 
of $\Phi\Phi^\top$ becomes the computational bottleneck. 
Minimizing~\eqref{eq:ridge-regression} is this still possible in this case using 
iterative least-squares solvers, such as the method of conjugate gradients,
or straight-forward stochastic gradient descent~\cite{bottou2010large}. 
%

\begin{algorithm}[t] 
\begin{algorithmic}[1]
\INPUT image $x$, graph $G=(V,\mathcal{E})$
\medskip
\STATE  $\phi_{st}\in\mathbb{R}^{D}\ \ \leftarrow$ \ \ feature vector of edge $(s,t)\in \mathcal{E}$
\FOR{$j,k\in\mathcal{L}\!\times\!\mathcal{L}$}
\STATE $\theta_{st;jk} = \log f_{jk}(\phi_{st})\quad \forall(s,t)\in \mathcal{E}$
\ENDFOR
\STATE $E(y) = \sum_{(s,t)\in\mathcal{E},(j,k)\in\mathcal{L}\times\mathcal{L}} \theta_{st;jk}\ind{ y_s\!=\!j \wedge y_t\!=\!k}$
\medskip
\OUTPUT energy function $E(y)$ for image $x$
\end{algorithmic}
\caption{\METHOD{} -- Prediction (Loopy case)}\label{alg:LSCRF-prediction}
\end{algorithm}

\paragraph{Non-Linear Models.} 
From non-linearities in the regression functions we expect more flexible 
and better predictions. However, this applies only if enough data 
and computational resources are available to train them. 

In this work, we use \emph{gradient boosted regression trees} \cite{friedman2001greedy} 
as non-linear regressors. These have been shown to be strong 
predictors~\cite{caruana2006empirical}, while at the same time 
begin very fast to evaluate. The latter aspect is particularly relevant 
in the context of image segmentation, where many predictor evaluations 
are required for each image. 
%

Tree and forest classifiers have a long tradition in image 
segmentation, but typically for different purposes. 
They are used either to construction per-(super)pixel feature 
representations or to predict per-(super)pixel unary classifier 
scores~\cite{fulkerson2009class,shotton2009textonboost,schroff2008object}.
In both cases, however, it is still necessary to afterwards perform 
regular CRF learning.

\METHOD{}, on the other hand, uses the trees to directly predict the 
coefficients of the energy function, replacing the need for additional 
CRF training that would require probabilistic inference. 
To our knowledge, the only exist method following a similar setup are 
\emph{decision tree fields (DTFs)}~\cite{nowozin2011decision}.
These also learn the potentials of a random field using non-parametric 
decision trees. However, they differ significantly in their technical 
details: DTFs are trained using the pseudo-likelihood principle, and 
they parameterize the weights in a particular hierarchical way that is 
learned jointly with their values.
\METHOD{}, on the other hand, can be implemented using any 
out-of-the-box regression method, trees being just one particular 
choice. 

\subsection{Extension to loopy graphical models}\label{subsec:method-loopy}
It follows from the consistency of the maximum likelihood procedure 
that the construction we describe above leads to a consistent training 
algorithm for CRFs, as long as the underlying graphical model has tree 
topology. 
However, many interesting models in computer vision are loopy, in 
particular those used for image segmentation. 
In this section we describe how to construct an energy function 
in this situation that approximates the one one would obtain from 
(intractable) maximum likelihood learning, and that works well 
in practice. 

We first observe that the \METHOD{} training procedure 
(Algorithm~\ref{alg:LSCRF-training}) can be performed 
regardless of whether the underlying model is loopy or 
not, since it does not require probabilistic inference 
at training time. 
At test time, however, we should not simply apply the 
rules \eqref{eq:theta_f_stjk}, since the terms in the 
resulting energy would not balance in the correct way 
when the underlying graph has loops. 
Instead, we use the following construction (Algorithm~\ref{alg:LSCRF-prediction}): 
first, we decompose the (loopy) graph of a new image into a 
collection of subgraphs. 
This is inspired by tree-reweighted message passing~\cite{kolmogorov2006convergent}, where each subgraph 
is a tree. 
In our case, we choose the simplest possible trees, \ie, 
single edges with their two adjacent vertices.
For each such edge, $(s,t)$, we use the expressions for 
$\hat \theta_{s;j}(x)$ and $\hat \theta_{st;jk}(x)$ 
(Equation~\eqref{eq:theta_sj}) to obtain a partial energy function, 
\begin{align}
E^{(s,t)}_{\hat\theta(x)} &= \sum_{(j,k)\in\mathcal{L}\times\mathcal{L}} 
E^{(s,t)}_{\hat\theta(x);jk}\ind{y_s=j\wedge y_t=k},
\label{eq:energy-oneedge}
\intertext{with}
E^{(s,t)}_{\hat\theta(x);jk}
&=\hat \theta_{st;jk}(x)+ \hat \theta_{s;j}(x) + \hat \theta_{t;k}(x) \notag
\\
&=\log\frac{f_{st;jk}(x)}{f_{s;j}(x)f_{t;k}(x)}+ \log f_{s;j}(x)+ \log f_{t;k}(x) \notag
\\
&=\log f_{st;jk}(x).\label{eq:energy-oneedge-oneconfig}
\end{align}
Summing the expressions~\eqref{eq:energy-oneedge} over all edges we obtain 
a joint energy function, $E_{\hat\theta(x)}$ of all variables. It is 
determined completely by the pairwise regression functions that were learned 
from the training data. 

Further extensions of \METHOD{} are imaginable. For example, graphical models 
with higher order terms could be handled by the junction tree generalization 
of Equation~\eqref{eq:theta_sj}~(see \cite{wainwright2008graphical}).
However, we leave the realization of such extensions to future work.

%% file: relatedwork.tex
\section{Related work} \label{sec:relatedwork}
A large body of prior work exists on the topic of conditional random 
field training. 
In this section we discuss only some of the most related works. 
For a much broader discussion, see, \eg, the overview articles~\cite{domke2013learning,nowozin2011structured,wainwright2008graphical}.
We also discuss our contribution only in context of probabilistic CRF 
training. 
Maximum-margin learning, in particular structured support vector 
machines~(SSVMs)~\cite{tsochantaridis2005large} rely on different assumptions 
and optimization techniques. For a comparison see, \eg,~\cite{nowozin2010parameter}. 
So far few decomposition-based techniques 
exists for SSVMs. An example is~\cite{komodakis2011efficient}, 
which introduces a scheme of alternating between two steps: solving 
independent max-margin subproblems and updating Lagrange multipliers 
to enforce consistency between the solutions of the subproblems.
However, as a first-order dual decomposition technique, many iterations 
can be required until convergence, and experience in a large-scale
setting is so far missing. 
%

Traditional probabilistic CRF training aims at finding a weight vector, $w$, 
that maximizes the conditional likelihood of the training data under 
an assumed log-linear model for the conditional distribution, 
$P_w(y|x)\propto\exp(\,-E(x,y;w)\,)$ with $E(x,y;w)=\langle w,\phi(x,y)\rangle$. 
Equivalently, one minimizes the negative conditional log-likelihood,
\begin{align}
\ell(w) = \lambda\|w\|^2 - \sum_{i=1}^T E(x^i,y^i;w) + \log Z(x^i;w)
\label{eq:MCL}
\end{align}
where $\lambda\geq 0$ is a regularization constant and 
$Z(x^i;w) = \sum_{y\in\mathcal{Y}} \exp( -E(x^i,y^i;w))$ 
is the \emph{partition function}. 

This optimization problem is smooth and convex so --in principle-- 
gradient based optimization techniques, such as steepest descent, 
are applicable.
In practice, however, the exponential size of $\mathcal{Y}$ make 
this intractable, except in a few well-understood cases, such as model 
of very low tree width. 
Otherwise, computing $Z$ or its derivatives is $\#P$-hard~\cite{bulatov2005complexity}, 
so already computing a single exact gradient of \eqref{eq:MCL} is 
computationally intractable. 

For tree-shaped models, the gradients can be computed in 
polynomial time. Nevertheless, minimizing~\eqref{eq:MCL} exactly
is possible in practice only when the number of training images 
is small, since every gradient evaluation requires probabilistic 
inference across all training examples. 
\emph{Stochastic gradient descent (SGD)} training has been proposed
to overcome this~\cite{bottou2010large,vishwanathan2006accelerated},
but since it still requires probabilistic inference in the underlying 
graphical model its use is limited to small and tree-shaped models. 

For larger or loopy models, approximate inference methods have 
been proposed that do not approximate the gradients of~\eqref{eq:MCL}, 
but replace the whole objective by an easier one. 
Prominent examples are pseudolikelihood (PL)~\cite{besag1975statistical,nowozin2011decision} and piecewise 
(PW) training~\cite{shotton2009textonboost,sutton2005piecewise}. 
These methods replace the log-likelihood, \eqref{eq:MCL}, by a 
surrogate that allows easier minimization, \eg, by bounding the 
intractable partition function, $Z$, by a factorized approximation. 

\METHOD{} shares several properties with PL and PW training methods, 
in particular the fact that can be phrased as solving a set of independent 
optimization problems and does not require probabilistic inferences at 
training time. 
%
%
However, it has some additional desirable properties shared by neither of 
the earlier techniques: in particular, only \METHOD{} provides a closed 
form solution for the coefficient of the energy function. Also, both PL 
and PW are typically derived for log-linear conditional distributions, 
whereas \METHOD{} make no assumption on the parametric form of the predictors,
which makes it easy to train also non-linear models.

%% file: experiments.tex
\section{Large-Scale Image Segmentation}\label{sec:experiments}
Our main interest in the development of \METHOD{} is the problem 
of large-scale image segmentation.
The input data are RGB images of variable sizes and the goal 
is to predict a segmentation mask of the same size in which each 
pixels or superpixel is assigned one out of $r$ values. 
In the case of \emph{multi-class semantic segmentation}, these labels correspond 
to semantic classes, which can be 'stuff', such 
as \emph{road}, or \emph{sky}, or 'things', such 
as \emph{cars} or \emph{people}. 
The number of labels, $r$, is typically between 5 and 20 
in this case. 
A second case of interest is \emph{figure--ground segmentation}, 
where $r=2$ and the labels indicate \emph{foreground} and \emph{background}.

\subsection{Datasets}
For our experiments on the multi-class semantic situation, we use 
the 8-label \emph{Stanford background} dataset~\cite{gould2009decomposing}, 
which has been created by merging images of urban and rural scenes from 
four existing image datasets: LabelMe~\cite{russell2008labelme}, 
MSRC~\cite{shotton2009textonboost}, PASCAL VOC~\cite{everingham2010pascal} 
and Geometric Context~\cite{hoiem2007recovering}.
Despite this effort, the dataset contains only 715 images, so 
we do not consider it a large-scale dataset by current standards. 
In fact, none of the current datasets for natural image segmentation 
contains significantly more than a thousand images. 
The main reason is that providing pixelwise ground truth annotation 
is time consuming and therefore costly, even when performed on a
crowd sourcing platform like Amazon Mechanical Turk.

To overcome this limitation, we make a second contribution in this 
manuscript: two large-scale datasets for figure--ground image segmentation, 
\emph{HorseSeg} with over 25,000 images and \emph{DogSeg} with over 156,000 
images. 
The images for both datasets were collected from the ImageNet 
dataset~\cite{deng2009imagenet} and the \emph{trainval} part 
of PASCAL VOC2012~\cite{everingham2010pascal}.
As test set, we use 241 horse images and 306 dog images, 
which were annotated manually with figure-ground segmentations. 

All images from the PASCAL dataset have manually created ground truth
annotation. 
For the remaining images from ImageNet, only manual annotation by 
bounding boxes or per-image labels is available. 
As such, the dataset provides a natural benchmark for large-scale, 
semi-supervised, image segmentation with different amount of ground 
truth.

To enable experiments on large-scale supervised training, we also 
provide semi-automatically created figure-ground annotation for 
all images of the two datasets, based on the following procedure.
We first group the images by whether they have bounding box annotation 
for the foreground object available or not. 
For the annotated part (approximately 6,000 of the \emph{horse} images 
and 42,000 of the \emph{dog} images), we apply the \emph{segmentation 
transfer} method of~\cite{KuettelImagenet} to the bounding box region. 
A visual inspection of the output (see Figure~\ref{fig:examples}) 
shows that the resulting segmentations are often of very high accuracy, 
so using them as a proxy for manual annotation seems promising.
For the remaining images, only per-image annotation of the class label
is known. To these images we apply the unconstrained segmentation 
transfer, which also yields figure--ground segmentation mask, but 
of mixed quality (see Figure~\ref{fig:examples}).
In Section~\ref{sec:experiments} we report on experiments that use 
these different subsets for training CRF models.
Note that even if the training data is generated by an algorithm, 
the evaluation is performed purely on manually created ground 
truth data, so the evaluation procedure is not biased towards
the label transfer method.

\subsection{Image Representations}
We represent each image by a graph of SLIC superpixels~\cite{achanta2012slic}
with an edge for any touching pair of superpixels.
For each superpixel, $s$, we compute the following features:
\begin{itemize}
    \item $\phi_s^{\text{color}}\!\in\![0,255]^{3}$: average RGB color in $s$,
    \item $\phi_s^{\text{pos}}\!\in\![0, 1]^2$: center of $s$ in relative image coordinates, 
    \item $\phi_s^{\text{sift}}\!\in\![0, 16]^{128}$: rootSIFT descriptor~\cite{Arandjelovic12} at center of $s$.
    \item $\phi_s^{\text{tree}}\!\in\![0, 1]^{8}\ /\ [0, 1]^{10}$: predicted label probabilities. 
\end{itemize}
The $\phi_s^{\text{tree}}$ features are used for the linear regression and pseudolikelihood experiments. 
They consist of the outputs of per-label training boosted tree classifiers on a subset 
of the available training data.
On the Stanford dataset this results in 8 additional feature dimensions.
For DogSeg and HorseSeg, we use the Stanford features plus the output of 
per-superpixel classifiers of the foreground and background classes, 
resulting in overall 10 additional dimensions.
For each edge $(s,t)$ we define a feature representation by concatenating the features 
of the contributing superpixels, $\phi_{st} = [\phi_s,\phi_t]$.

%

\subsection{Implementation}
Training a CRF by \METHOD{} requires only solving multiple regression problems, 
a task for which several efficient software packages are readily available. 
In our experiments with linear regression, we use the \emph{Vowpal Wabbit}
package\footnote{\url{http://hunch.net/~vw/}} (LBFGS optimization, learning rate 0.5, 
no explicit regularization), clamping the predictions to the interval $[10^{-9},1]$. 
%
Vowpal Wabbit is particularly suitable for large scale learning, since it supports 
a variety of infrastructures, from single CPUs to distributed compute clusters.
%
As non-linear setup we train gradient boosted regression trees using the MatrixNet 
package~\cite{trofimov2012using} with default parameters (500 oblivious trees, 
depth 6, learning rate 0.1).

As baselines, we use CRFs with only unary terms, trained with the same 
regression methods as \METHOD{}. 
We furthermore implemented a baseline that performs segmentation with 
only unary terms trained in the usual way with a logistic loss, 
and a CRF with pairwise terms using pseudolikelihood and piecewise training. Latter
methods rely on the~\emph{grante} library\footnote{\url{http://www.nowozin.net/sebastian/grante/}}
and \emph{Vowpal Wabbit} package. 
Other training techniques, in particular those requiring probabilistic inference, 
are not includes, since they do not scale to the size of datasets we are interested in. 

At test time, we use MAP prediction to create segmentations of the test images. 
For this, we need to minimize the energy function that the training methods have produced. 
In the unary-only case, this is possible on a pixel-by-pixel level. 
For energy function with pairwise terms (\METHOD{} and pseudolikelihood), we use 
MQPBO~\cite{kohli2008partial} followed by TRWS~\cite{kolmogorov2006convergent}
for the figure-ground segmentations. For both we use the implementations provided 
by the \emph{OpenGM2} library~\footnote{\url{http://hci.iwr.uni-heidelberg.de/opengm2/}}. 

Alternatively, we could use solvers based on integer linear programming, which 
have recently been found effective for image segmentation tasks~\cite{kappes2013benchmark}. 
We plan to do so in future work.

\subsection{Results}\label{subsec:results}
We first report experiments on the multi-label Stanford 
Background dataset. 
While this is not a large-scale setup, the purpose is to show 
that \METHOD{} achieves segmentation accuracy comparable to 
existing techniques for approximate CRF training, in particular
as least as good as pseudolikelihood training, which is commonly 
used for this kind of problems.
Afterwards, we report results in the large-scale regime, 
using the DogSeg and HorseSeg dataset with semi-automatic
annotation.

\paragraph{Stanford Background Dataset.} 
Table~\ref{table:stanford} summarizes the results 
on the Stanford Background dataset in numeric form. 
For example segmentations, please see the supplemental material.
We compare seven setups: \METHOD{} (unary-only or pairwise energies) 
with linear or non-linear parameterization, logistic regression 
(unary-only), pseudolikelihood and piecewise (pairwise only).

The results show that for learning a segmentation model with only unary 
terms, the squared loss objective of \METHOD{} achieves comparable result
to usual probabilistic loss (logistic regression).
Including pairwise terms into the model improves  the segmentation 
accuracy when we train the model with piecewise method or with \METHOD{}.
Pairwise terms trained  by pseudolikelihood training have only
a minor positive effect on the segmentation quality in our experiments. 
The non-linear and the linear versions of \METHOD{} achieve comparable 
performance, likely because the amount of training data per label or 
label pair is small, so the classifier is not able to make use of the 
additional flexibility of a non-linear decision function.
In fact, approximately 20 of the 64 possible label combination occur 
too rarely in the training images to train predictors for them. 
Instead, we assigned them constant probabilities of $10^{-3}$.

Note that the numbers for pairwise models are also roughly comparable 
to the 74.3\% average per pixel accuracy reported in the original 
work~\cite{gould2009decomposing} for a CRF with pairwise term. 
We attribute the existing difference in values to our use of 
simpler (and faster computable) image features. 


\begin{table}
\centering
\begin{tabular}{|c|c|c|}
\hline
Model
 & unary& pairwise \\
\hline
\METHOD{} linear & 70.3 (1.0) & \textbf{73.4} (1.1) \\
\METHOD{} non-linear & 70.9 (1.1) &  \textbf{73.3 } (1.1) \\
logistic regression & 70.4 (0.8)&  -- \\
pseudolikelihood & -- &  71.0  (1.1) \\
piecewise & -- &  72.9  (0.9) \\
\hline
\end{tabular}
\caption{Segmentation quality (average per pixel accuracy in \%) on \emph{Stanford background} 
dataset. Numbers in brackets are standard deviations over five cross-validation splits.
}
\label{table:stanford}
\end{table}

\begin{table}
\centering
\subfigure[images with manual annotation]{
\begin{tabular}{|c|cc|cc|}
\hline
& \multicolumn{2}{c|}{HorseSeg [147]} & \multicolumn{2}{c|}{DogSeg [249]} \\
\hline
Model & unary & pairwise & unary & pairwise \\
\hline
\multirow{2}{*}{linear} & 81.4 & \textbf{82.2} & 77.8 & \textbf{79.1} \\
                        &  (\textless 1m)  & (\textless 1m) & (2.5m) & (2.5m) \\
\hline
\multirow{2}{*}{non-linear} & 81.6  & \textbf{83.8} & 78.5 & \textbf{80.8}  \\
                            & (\textless 1m) & (1.5m) & (\textless 1m) & (2m) \\
\hline
\end{tabular}}

\subfigure[images with manual or object bounding box annotation]{
\centering
\begin{tabular}{|c|cc|cc|}
\hline
& \multicolumn{2}{c|}{HorseSeg [6044]}& \multicolumn{2}{c|}{DogSeg [42763]} \\
\hline
Model & unary & pairwise & unary & pairwise \\
\hline
\multirow{2}{*}{linear} & 82.0 & \textbf{83.6 } & 78.5 & \textbf{81.2} \\
                        & (\textless 1m) & (\textless 1m) & (7m) & (14m) \\
\hline
\multirow{2}{*}{non-linear} & 83.6 & \textbf{86.4} & 80.9 & \textbf{83.8} \\
                            & (9m) & (32m) & (110m)  & (348m) \\
\hline
\end{tabular}}
\subfigure[all images]{
\centering
\begin{tabular}{|c|cc|cc|}
\hline

& \multicolumn{2}{c|}{HorseSeg [24837]} & \multicolumn{2}{c|}{DogSeg [156062]} \\
\hline
Model & unary & pairwise & unary & pairwise \\
\hline
\multirow{2}{*}{linear} & 81.4 & \textbf{83.3 } &   78.1 & \textbf{80.2} \\
                        & (1m) & (2m)           &  (20m) & (46m) \\
\hline
\multirow{2}{*}{non-linear} & 82.5 & \textbf{84.5} & 80.0 & \textbf{82.2} \\
                            & (88m)  & (354m) & (519m) & (668m) \\ 
\hline
\end{tabular}}
\caption{Results of \METHOD{} for figure-ground segmentation 
on HorseSeg/DogSeg (average per class accuracy in \%) for different training subsets. 
The numbers in brackets incidate training time 
and numbers in rectangular brackets indicate the number of training images.
}
\label{tab:results-doghorses}
\end{table}

\paragraph{DogSeg/HorseSeg dataset.}
In the large scale regime, we are interested in three questions.
Can we make truly large scale CRF training feasible? 
How do CRFs benefit from the availability of large amounts of 
training data? 
How useful is annotation that was created semi-automatically?

We study these question in three sets of experiments, using 
different subsets of the training data: (a) only images with 
manually created annotation, (b) images with annotation 
created manually or by segmentation transfer using bounding 
box information, (c) all training images. 

We train CRFs with pairwise terms using the linear and non-linear 
variant of \METHOD{}, and compare their segmentation performance 
to models with only unary terms. 
In situation (a), we use the feature vectors of all available 
superpixels to train the unary-only models, and all neighboring 
superpixel pairs to train \METHOD{} with pairwise terms. 
In the larger setup, (b) and (c), we reduce the redundancy in 
the data by using only 25\% of all superpixels for the unary-only
models, sampled in a class-balanced way.  
For pairwise models, we record the ratio of pairs with \emph{same label} 
versus with \emph{different label}.
Preserving this ratio, we sample 10\% of all superpixels pairs, 
in a way that combinations with both \emph{foreground} and both 
\emph{background} are equally likely, and also \emph{foreground/background} 
and \emph{background/foreground} transitions are equally likely.
The percentages are chosen such that the training problems in both 
situations are of comparable size. 
On the \emph{DogSeg} dataset, they consists of approximately 90K data 
points for situation (a), 3.5M data points for (b), and 13M data 
points for (c), except the pairwise/nonlinear case, where we use 
only 6.5M data points for memory reasons.
On the \emph{HorseSeg}, the number are roughly half as 
big for (a), and one sixth for (b) and (c).

A first observation is that using \METHOD{} training is computationally 
feasible even in the largest setup: for example, training with all 
training images in the \emph{DogSeg} dataset with linear objective 
with pairwise terms required 45 minutes on a 24 core workstation, 
compared to 20 minutes, if only unary terms are learned.
The non-linear setup took approximately 9 hours to train an 
energy function with only unary terms, and 11 hours when pairwise
terms were used. 

Table~\ref{tab:results-doghorses} shows numeric results for 
the segmentation accuracy and training time of the different setups. 
Example segmentation are provided in the supplemental material. 
The results allow us to make several observations that we believe 
will generalize beyond the specific setup of our work.

First, it has been observed previously that pairwise terms often 
have only a minor positive effect on the segmentation quality (\eg~\cite{fulkerson2009class}). 
Our experiments show a similar trend when a linear representation 
was used and the number of training examples was small. 
However, when a large training set was used, the difference 
between unary-only and pairwise models increased.
Second, the use of non-linear predictors consistently improved 
the segmentation quality. 
This could be a useful insight also for other CRF training methods, 
which rely predominantly on linearly parameterized 
energy functions.
Third, the segmentation quality improved significantly 
when increasing the number of training images, even though the 
additional images had only annotation created automatically 
using information from bounding boxes or per-image labels.
This indicates that segmentation transfer followed by large-scale 
CRF learning could be a promising way for leveraging the large 
amounts of unlabeled image data, \eg in the Internet. 

One can also see in Table~\ref{tab:results-doghorses} that the segmentation 
accuracy was highest when training on the manually annotated image set 
together with images that had bounding boxes annotation. Including
also the remaining images led to a reduction in the quality. 
This raises the question whether the annotation created only from 
per-image labels contain useful information at all. 
We performed additional experiments for this, measuring the segmentation
quality of training linear \METHOD{} on the HorseSeg dataset, but training
exclusively on images with annotation from bounding boxes, or on images 
with annotation from per-image labels. 
This resulting per-class accuracies are 82.0 (unary) and 83.9 (pairwise) 
for the bounding box case, and 81.1 (unary) and 82.8 (pairwise) for the
per-image case. Comparing this to the values 81.4 (unary) and 82.2 (pairwise)
from Table~\ref{tab:results-doghorses}, we see that the automatically 
generated segmentations do indeed contain useful information. Training only
on these achieves results comparable to training on the (admittedly much fewer) 
manually annotated images.

%% file: summary.tex
\section{Summary}\label{sec:summary}
\noindent In this work we make two main contributions: 
\begin{itemize}
\item a new technique for inference-free CRF training 
that scales to very large training sets, 
\item two new benchmark datasets of over 180,000 images and segmentation masks. 
\end{itemize}

We used these to perform the first truly large-scale experiments in the area 
of (semantic) image segmentation, which provided us with several noteworthy 
observations: 1) the positive effect of pairwise terms increased with 
the number of training examples, 2) training CRFs with non-linear energies 
is feasible and results in better segmentation models, 3) semi-supervised 
learning is practical and useful also for image segmentation, 
by (semi-)automatically generation annotation for otherwise unlabeled images. 

Furthermore, we are convinced that both contributions will be useful also 
outside of the area of image segmentation. The large-scale CRF training 
method is applicable regardless of the application area, and the dataset 
can serve also as a generic testbed for large scale training of loopy CRF models.



%% file: supplement.tex
\section{Supplementary material}

\begin{figure}[!h]
    \centering
    \onecolumn\includegraphics[width=.80\textwidth]{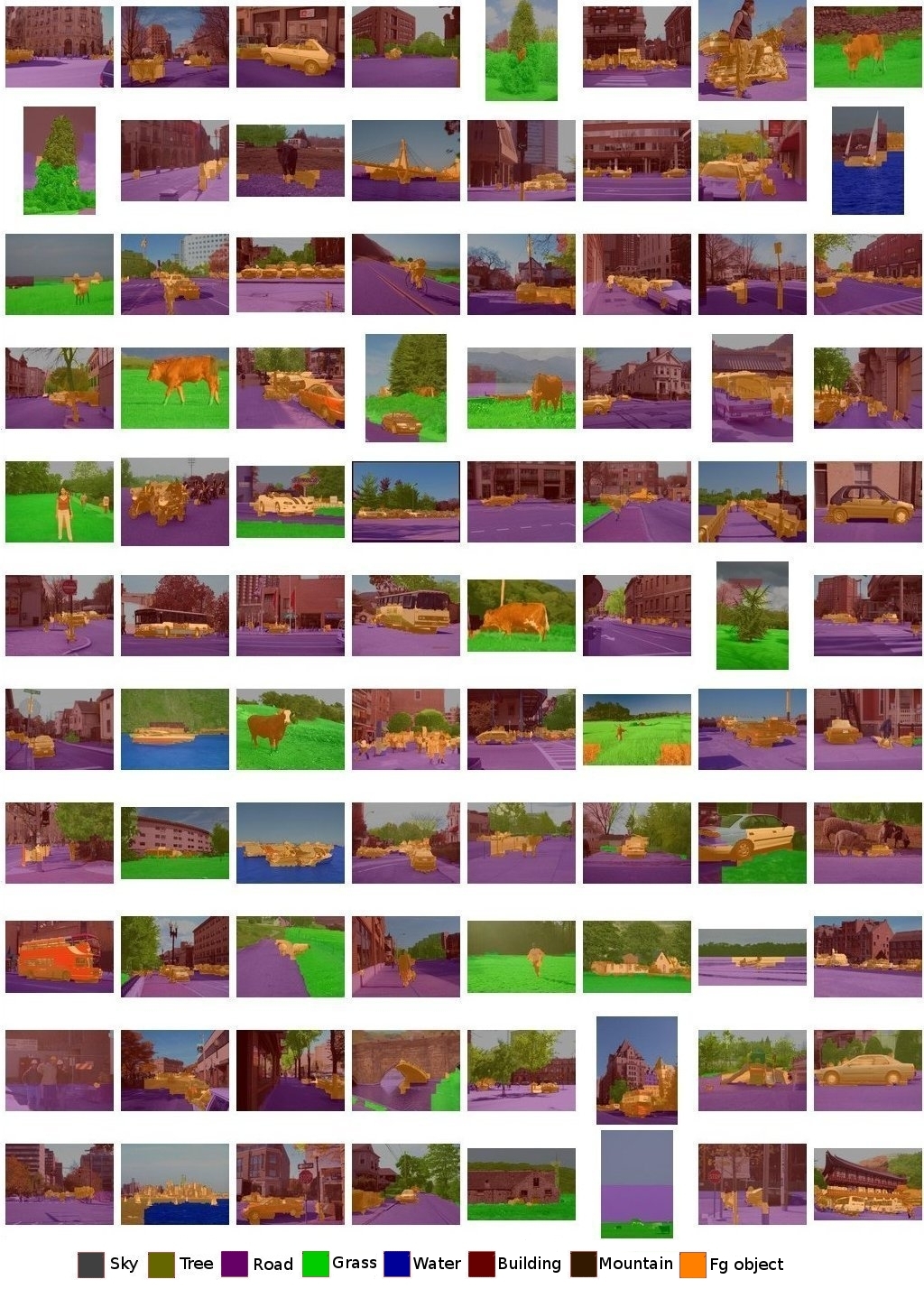}
    \caption{Image segmentation examples from Stanford background dataset for pairwise non-linear model.}
    \label{fig:segmentations-stanford}
\end{figure}

\begin{figure*}[!h]
    \centering
    \includegraphics[width=.80\textwidth]{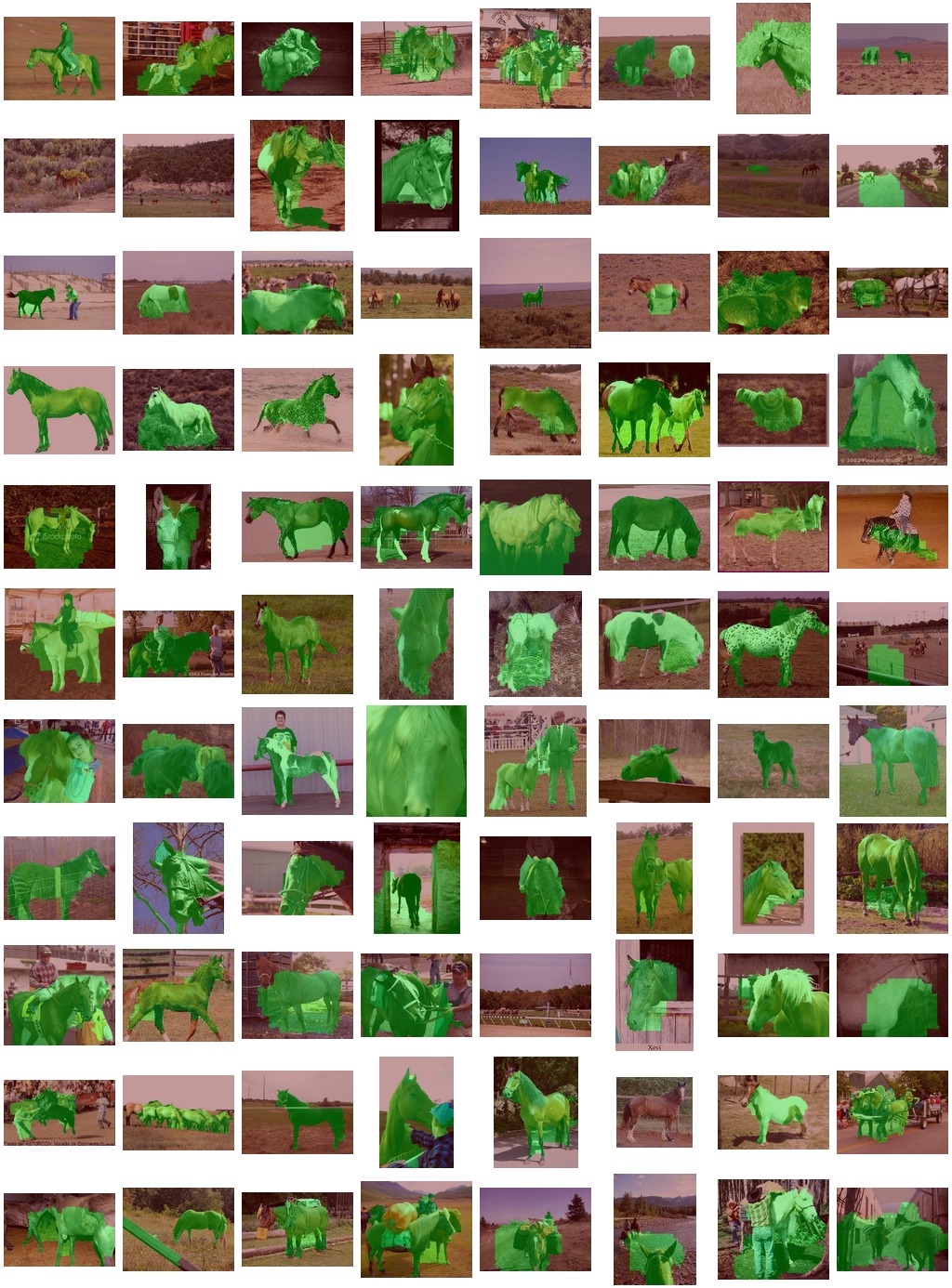}
    \caption{Test image segmentation examples from HorseSeg dataset for pairwise non-linear model 
             trained on images with manual annotation or object bounding box (green color means foreground, red --- background).}
    \label{fig:segmentations-horse}
\end{figure*}
\begin{figure*}[!h]
    \centering
    \includegraphics[width=.80\textwidth]{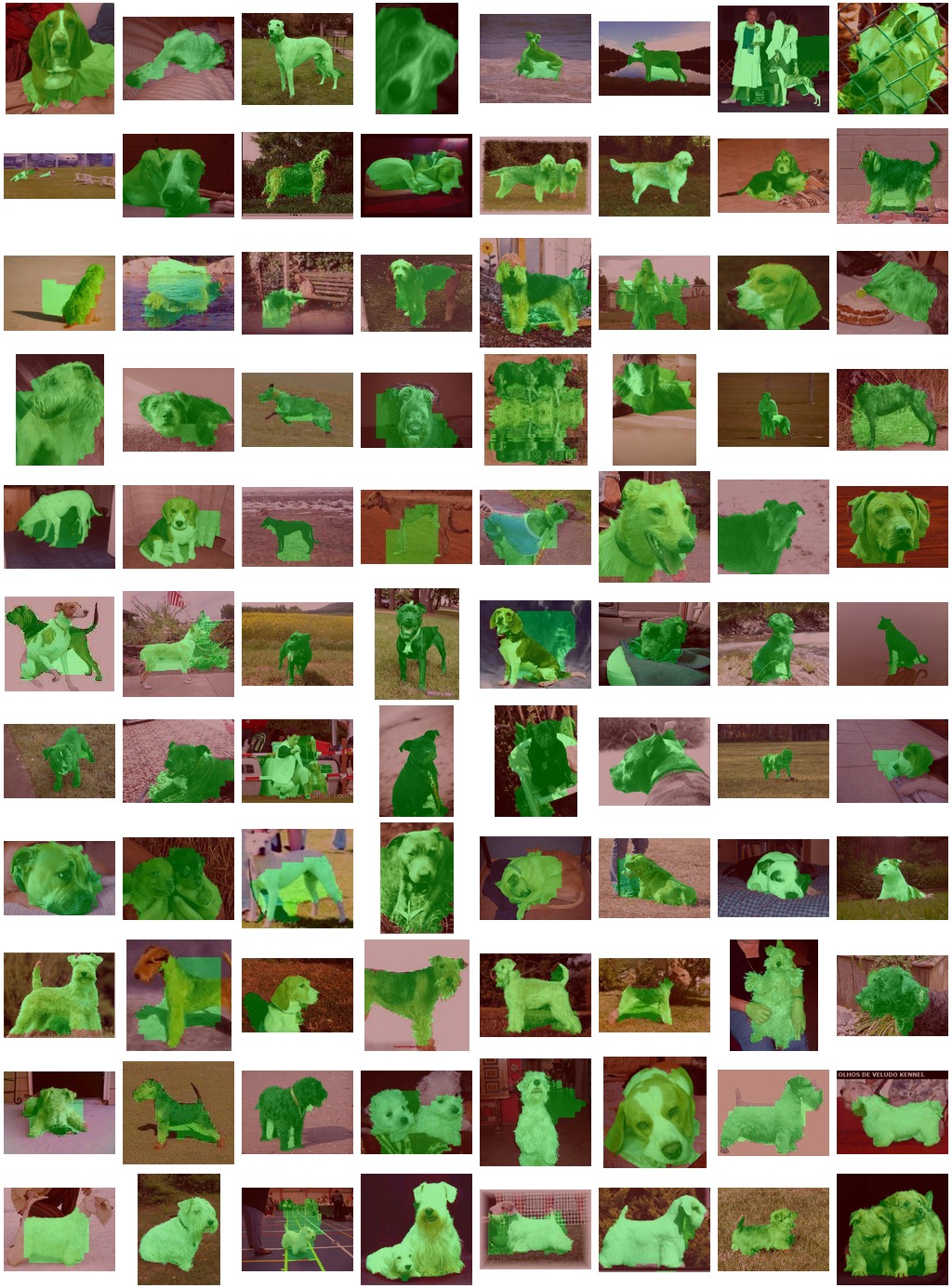}
    \caption{Test image segmentation examples from DogSeg dataset for pairwise non-linear model 
             trained on images with manual annotation or object bounding box (green color means foreground, red --- background).}
    \label{fig:segmentations-dog}
\end{figure*}

\begin{figure*}[!h]
\centering
\fbox{\subfigure[images with manual annotation]{
        \includegraphics[width=.80\textwidth]{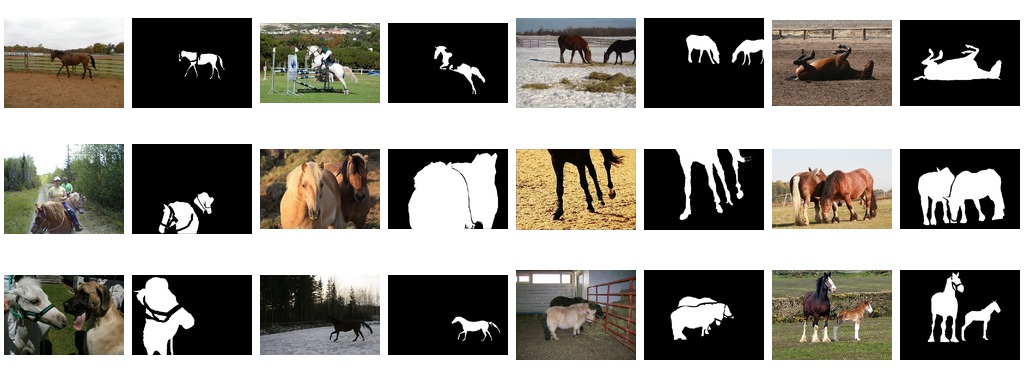}}
}
\fbox{\subfigure[images with object bounding box]{
        \includegraphics[width=.80\textwidth]{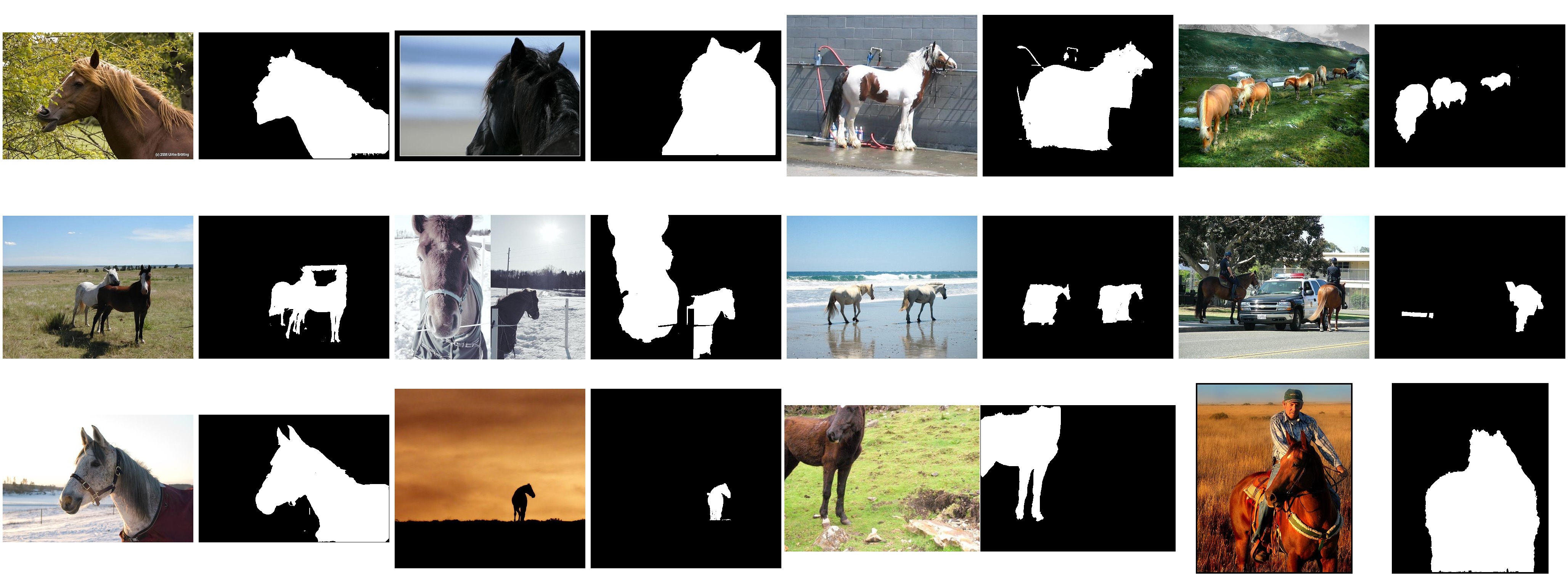}}
}
\fbox{\subfigure[images with only label]{
        \includegraphics[width=.80\textwidth]{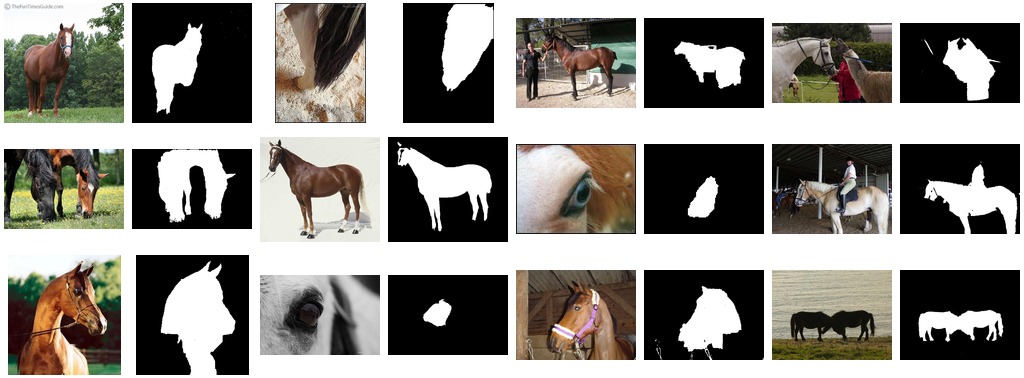}}
}
\caption{HorseSeg training image examples.}
\label{fig:horseseg-examples}
\end{figure*}

\begin{figure*}[!h]
\centering
\fbox{\subfigure[images with manual annotation]{
        \includegraphics[width=.80\textwidth]{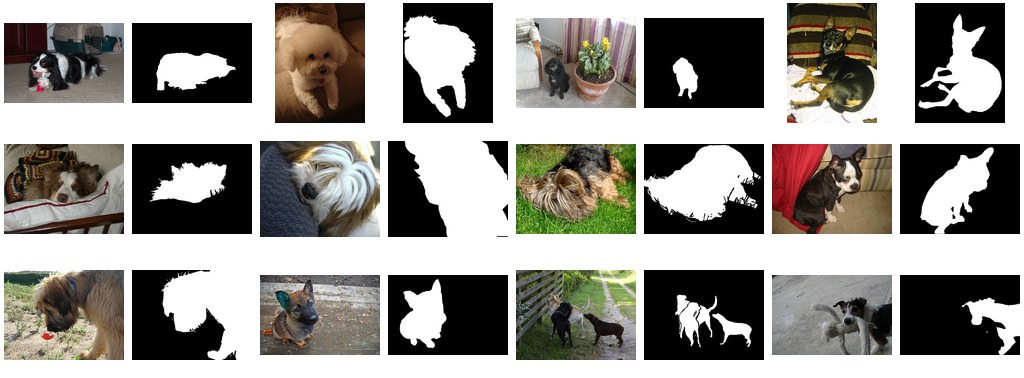}}
}
\fbox{\subfigure[images with object bounding box]{
        \includegraphics[width=.80\textwidth]{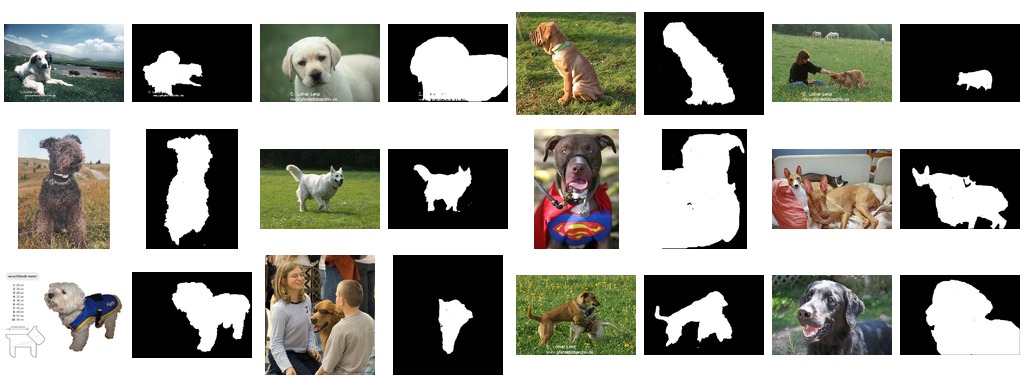}}
}
\fbox{\subfigure[images with only label]{
        \includegraphics[width=.80\textwidth]{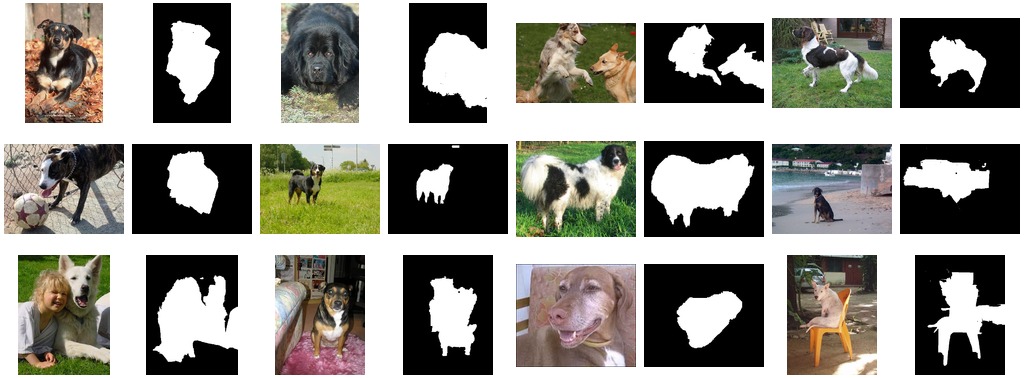}}
}
\caption{DogSeg training image examples.}
\label{fig:dogseg-examples}
\end{figure*}